\documentclass[preprint,review, compress,12pt]{elsarticle}

\usepackage{amssymb}

\usepackage{mathrsfs}
\usepackage{amsmath}
\usepackage{booktabs}
\usepackage{multirow}
\usepackage{makecell}

\usepackage{hyperref}
\usepackage[hyphenbreaks]{breakurl}

\usepackage{color}

\journal{Neurocomputing}

\begin{document}

\begin{frontmatter}



\title{From Text to Mask: Localizing Entities Using the Attention of Text-to-Image Diffusion Models}


\author{Changming Xiao\corref{cor1}\fnref{label1}}
\author{Qi Yang\corref{cor1}\fnref{label1}}
\author{Feng Zhou\fnref{label2}}
\author{Changshui Zhang\corref{cor2}\fnref{label1}}
\ead{zcs@mail.tsinghua.edu.cn}

\affiliation[label1]{organization={Institute for Artificial Intelligence, Tsinghua University (THUAI); Beijing National Research Center for Information Science and Technology (BNRist); Department of Automation, Tsinghua University},
            postcode={100084}, 
            state={Beijing},
            country={P.R. China}}
\affiliation[label2]{organization={Algorithm Research, Aibee Inc.},
            state={Beijing},
            country={P.R. China}}
\cortext[cor1]{Indicates equal contribution.}
\cortext[cor2]{Corresponding author.}

\begin{abstract}
Diffusion models have revolted the field of text-to-image generation recently.
The unique way of fusing text and image information contributes to their remarkable capability of generating highly text-related images. 
From another perspective, these generative models imply clues about the precise correlation between words and pixels.
This work proposes a simple but effective method to utilize the attention mechanism in the denoising network of text-to-image diffusion models.
Without additional training time nor inference-time optimization, the semantic grounding of phrases can be attained directly.
We evaluate our method on Pascal VOC 2012 and Microsoft COCO 2014 under weakly-supervised semantic segmentation setting and our method achieves superior performance to prior methods.
In addition, the acquired word-pixel correlation is generalizable for the learned text embedding of customized generation methods, requiring only a few modifications.
To validate our discovery, we introduce a new practical task called ``personalized referring image segmentation" with a new dataset.
Experiments in various situations demonstrate the advantages of our method compared to strong baselines on this task.
In summary, our work reveals a novel way to extract the rich multi-modal knowledge hidden in diffusion models for segmentation.

\end{abstract}


\end{frontmatter}


\section{Introduction}
\label{sec:intro}

Dense image prediction is a long-established research field that aims at producing pixel-level labels for given images~\cite{FCN}. 
It is the foundation of many applications in fields such as biomedicine~\cite{medical_seg_neuro}, robotics~\cite{affordancenet}, and surveillance~\cite{anomalydetect}.
To attain precise masks, the dense image prediction task usually requires expensive dense annotations for training. 
Although recent works~\cite{clims, CLIP-ES} have shown impressive results when training without pixel-wise labels, deliberate designs of the model architecture or the optimization objective are required.
With the development of powerful foundation models trained with internet-scale data, studies have emerged on how to mine valuable information from these off-the-shelf models for diverse tasks.
Regarding segmentation, some methods like~\cite{denseclip} have utilized the learned localization information from text-image discriminative models~\cite{clip} to reduce reliance on segmentation-specific schemes.
Inspired by a renowned quote from Feynman: \emph{What I cannot create, I do not understand}, we believe that generative models, the counterpart of discriminative models, should also have a thorough comprehension of images.

\begin{figure}
  \centering
  \includegraphics[width=\linewidth]{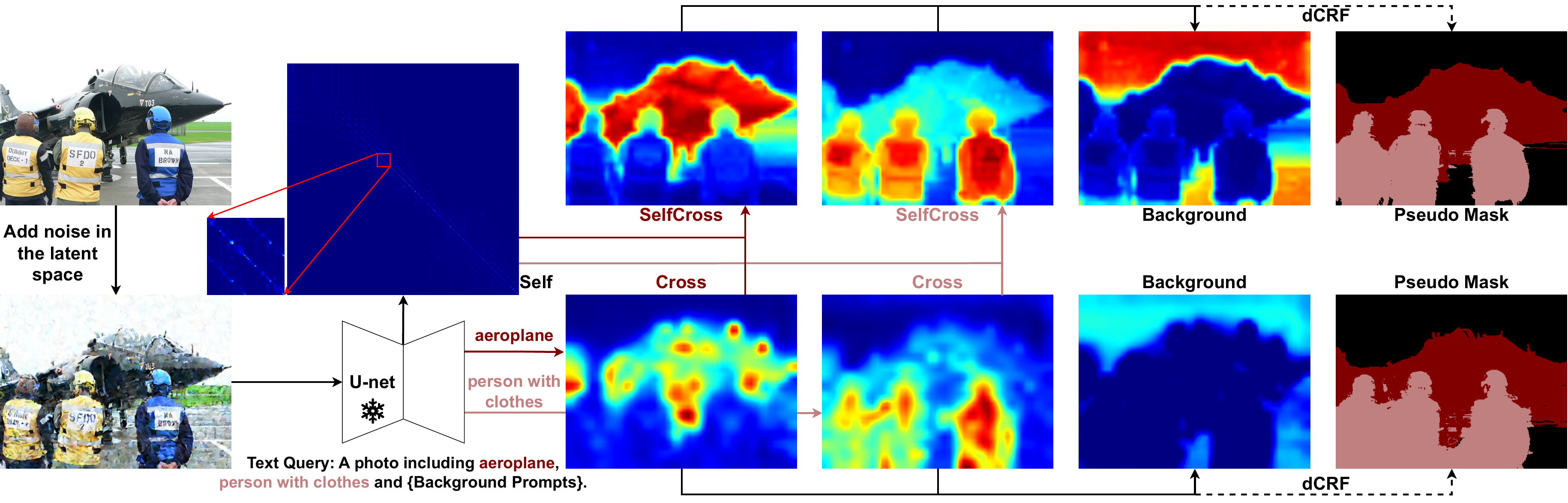}
  \caption{
  An overview of our proposed framework.
  We first add noise to the latent and then input it into the denoising U-net with specially designed text queries.
  Next, we combine cross-attention and self-attention in the model to obtain the correlation map between words and pixels.
  After comparing different correlation maps and post-processing with dense CRF~\cite{dCRF}, we attain pseudo masks at last.
  Best viewed in color.
  }
  \label{fig:outline}
\end{figure}

Diffusion models~\cite{score, ddpm} have opened a new era of generative models, and their multi-modal variants~\cite{stablediffusion} trained on billions of image-caption pairs have revolutionized the field of text-to-image synthesis.
When diffusion models generate images from texts, the association between words and different spatial regions can be leveraged to indicate localization information.
We propose a simple but effective way to distill this information and use it for dense image prediction.
Building off a recently open-sourced text-to-image diffusion model~\cite{stablediffusion}, our method exhibits remarkable semantic segmentation capability.
Figure~\ref{fig:outline} shows an overview of our framework, which effectively leverages cross-attention and self-attention in the diffusion model to obtain the segmentation map.
Compared to previous works, our proposed method doesn't need segmentation-specific re-training~\cite{ddpmseg}, module addition~\cite{ODISE}, or inference time optimization~\cite{peekaboo}, and the whole pipeline is accomplished with no exposure to pixel-wise labels~\cite{SEEM}.

Leveraging pre-trained text-to-image diffusion models brings more benefits.
It has been shown that generative models have a better spatial and relational understanding than discriminative models~\cite{ODISE}, which is crucial for segmentation.
Besides, internet-scale training data endows the model with the ability to handle open-vocabulary scenarios.
Furthermore, customized generation with text-to-image diffusion models has been proven feasible~\cite{customdiffusion}.
These methods typically map the subject to an identifier word, and novel renditions of it can be obtained by inserting the word into various descriptions.
We integrate this technique with our method and locate user-specific items by exploiting the correspondence between learned word embeddings of personalized concepts and image segments.
It is worth noting that we can compose customized instances and textual contexts into multi-modal segmentation queries, which was hard to achieve with previous methods~\cite{PALAVRA}.

To investigate the performance of different methods in this customized case, we introduce a new task called ``personalized referring image segmentation" with a new dataset named \textbf{Mug19}.
This task aims to locate user-specific entities and has many possible practical application scenarios, such as those for household robots.
The dataset is collected in a laboratory scenario and is conscientiously devised to ensure that in most cases, uni-modal information is insufficient to locate the proper object.
This setup makes it a more pragmatic and user-friendly task and allows this new benchmark to assess the multi-modal comprehension ability of different models.

We conduct experiments from different aspects to validate the effectiveness of our method.
We first evaluate the weakly-supervised semantic segmentation ability of our approach on classic datasets: Pascal VOC 2012~\cite{pascalvoc} and MS COCO 2014~\cite{mscoco}.
Then we demonstrate that we can locate personalized embeddings just like locating category embeddings, showcasing the generality of our framework.
We further reveal that traditional approaches using uni-modal information have difficulties in dealing with our proposed task.
Finally, ablation studies are conducted to validate the intuition of our framework designs.

In summary, the contributions of this work are as follows:
\begin{itemize}
    \item A novel plug-and-play method for open-vocabulary segmentation is proposed, which utilizes the attention mechanism in off-the-shelf text-to-image diffusion models.
    \item A new benchmark named ``personalized referring image segmentation" is introduced, which is valuable for both industrial application and academic research.
    \item Experiment results on diverse segmentation tasks verify that our method can achieve state-of-the-art (SOTA) performance.
\end{itemize}

\section{Related Work}
\label{sec:related}

\subsection{Open-Vocabulary Segmentation}
\label{sec:openvoc}

Researchers have developed different technologies to make segmentation systems more practical.
One popular avenue is considering the weakly supervised setting and using image-level annotations which are easier to obtain~\cite{EraseTG, Li2023WeaklySS}.
But these solutions based on Class Activation Mapping (CAM)~\cite{cam} are limited to a predefined taxonomy.
Other researchers consider zero-shot learning for segmentation, which aims to transfer the knowledge extracted from seen categories to unseen ones.
With the development of word embedding~\cite{word2vec}, language generalization ability has been exploited to depict semantic associations between different classes~\cite{zs3}.

Recently, foundation models have made game-changing progress, and zero-shot abilities have been found to automatically emerge from them~\cite{clip}.
Thus open-vocabulary paradigm\footnote{Also known as open-set paradigm or open-world paradigm.}, a more general zero-shot formulation, has gradually replaced the classic paradigm~\cite{ovd}.
Contrastive Language-Image Pre-training (CLIP)~\cite{clip}, a discriminative foundation model that has learned vast image-text knowledge from the internet, has become a common component of open-vocabulary recognition methods~\cite{denseclip}.
Nevertheless, it has been discovered that the representation of CLIP is sub-optimal for segmentation tasks~\cite{ODISE}.
Therefore, we leverage a generative diffusion model instead~\cite{stablediffusion}, which is believed to have a thorough perception of scene-level structure~\cite{ODISE, daam}.

\subsection{Diffusion Model for Segmentation}
\label{sec:diffseg}

With the development of diffusion models, researchers have begun to pay attention to their potential in perception tasks.
In addition to global prediction tasks like classification~\cite{diffusionclass}, dense prediction tasks like segmentation are also gradually attracting attention~\cite{diffusionimplicitseg, ddpmseg, ODISE, grounddiff, diffumask}.
These methods can be roughly divided into $3$ groups.

One group regards the segmentation task as a generative process~\cite{diffusionimplicitseg}.
They learn the conditional denoising process from noise distribution to dense masks.
The input image serves as a condition to guide the sampling process, and dense annotations are required for training.
Instead of learning a different denoising process, another group exploits the knowledge of pre-trained diffusion models~\cite{ddpmseg, ODISE}.
They leverage diffusion features to train additional networks that provide dense predictions.
As pre-trained diffusion models possess powerful generative capability, their internal representation captures rich semantic information.
Thus labels used for training and parameters to be optimized can be greatly reduced, making their frameworks very efficient.
In order to further diminish dependence on pixel-wise labels, the last group operates on synthetic data~\cite{grounddiff, diffumask}.
They synthesize realistic images and segmentation masks simultaneously and use this automatically constructed dataset to train segmentation models.
The trained model exhibits competitive performance on real images to models trained on real data.
Nevertheless, due to the domain gap between synthetic data and real data, deliberate designs of data processing are required for good performance.

Compared to these segmentation methods that employ diffusion models, our method is more flexible and does not require dense annotations, segmentation-specific re-training, or complicated data processing designs.
Thus it only takes $1$ second to infer an image in our framework while some competing methods need minutes to segment one image.

\subsection{Composed Image Retrieval}
\label{sec:CIR}

Composed image retrieval is a multi-modal task aimed at retrieving images through image and text queries~\cite{CIR}.
Earlier studies have focused on synthesis data~\cite{CIR} and fashion products~\cite{FashionVLP}, while recent works have taken open-domain data into consideration~\cite{CIRR}.
From a methodological perspective, researchers have explored different ways to fuse multi-modal information.
They have proposed various mechanisms ranging from affine transformation~\cite{film} to cross-attention~\cite{transformer}.

Currently, with the development of vision-and-language pre-trained models~\cite{clip}, researchers have contemplated applying them to composed image retrieval~\cite{PALAVRA, CIRR, FashionVLP, pic2word}.
Among these approaches,~\cite{PALAVRA, pic2word} took up early-fusion and had freer reasoning capabilities than late-fusion methods~\cite{CIRR, FashionVLP}.
We adopt early-fusion in our proposed method, but we care more about localization ability, which is valuable for many practical applications.
Although PALAVRA~\cite{PALAVRA} also considered the segmentation task, its experimental results indicated its failure to distinguish between context and subject.
On the contrary, we design the localization task to rely on the comprehension of context and subject, and experimental results demonstrate the effectiveness of our method.

\section{Method}
\label{sec:method}

\subsection{Preliminary}
\label{sec:preliminary}

Diffusion models~\cite{score, ddpm} are a class of generative models which learn the data distribution by progressively denoising from a tractable noise distribution.
They can be interpreted as a sequence of time-conditional denoising auto-encoders.
To reduce resource consumption, Latent Diffusion Model~\cite{stablediffusion} is proposed which conducts the diffusion process in the latent space obtained by a perceptual compression model~\cite{vqgan}.
Additionally, in order to achieve flexible conditional generation, the denoising U-net backbone~\cite{unet} is usually augmented with the cross-attention and self-attention mechanism~\cite{transformer}.
With paired image-condition training data $(x, y)$, the optimization objective of denoising model $\epsilon_\theta$ is ordinarily simplified as:
\begin{equation}
    L_{L D M}:=\mathbb{E}_{\mathcal{E}(x), y, \epsilon \sim \mathcal{N}(0,1), t \sim \mathcal{U}[1,T]}\left\|\epsilon-\epsilon_\theta\left(z_t, t, y\right)\right\|^2,
\end{equation}
where $\mathcal{E}$ is the encoder of the compression model, $\epsilon$ is the sampled Gaussian noise added to the image, $t$ represents the current time step, $T$ is the total number of denoising steps, and $z_t$ is the noisy version of $\mathcal{E}(x)$.
More specifically, $z_t=\sqrt{\bar{\alpha}_t} \mathcal{E}(x)+\sqrt{1-\bar{\alpha}_t} \epsilon$, where $\bar{\alpha}_t \in (0,1)$ is the step size derived from a predefined variance schedule and decreases as $t$ increases~\cite{ddpm}.

We adopt an open-sourced text-conditional latent diffusion model named Stable Diffusion~\cite{stablediffusion} for this work, which is affordable to infer on consumer GPUs.
Its compression model encodes images into a low-dimensional latent space by sampling based on the predicted mean and standard deviation, similar to Variational Auto-Encoder (VAE)~\cite{vae}.
$y$ represents text in this case, and the model is trained on $5$ billion image-caption pairs~\cite{LAION5B}.
The language prompt is encoded by CLIP first and is then injected into the model through cross-attention layers.

\subsection{Attention Mechanism}
\label{sec:attention}

\begin{figure}[t]
  \centering
  \includegraphics[width=\linewidth]{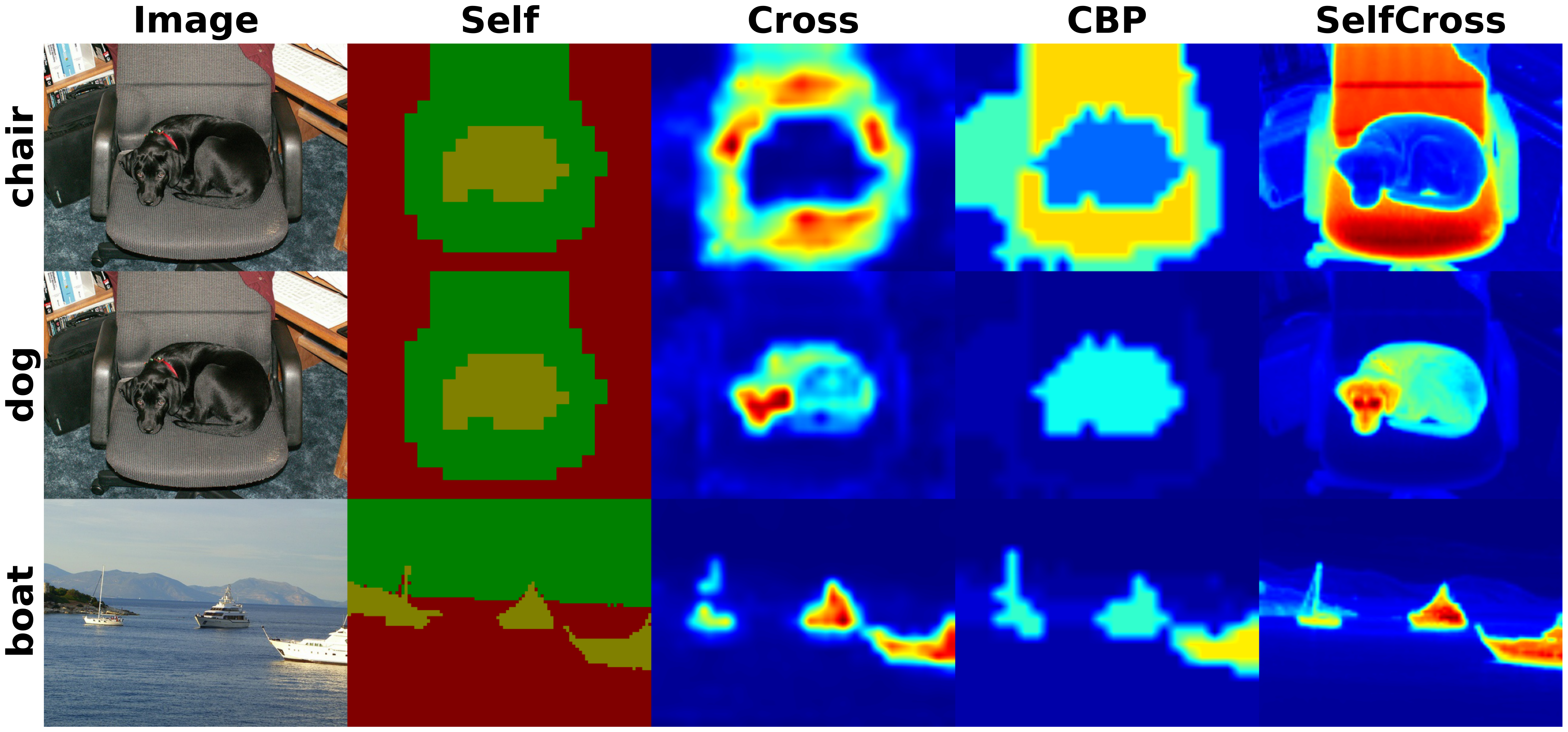}
  \caption{
  Visualization of correlation maps. 
  The texts on the left are the corresponding categories. 
  The $2$-nd column depicts the spectral clustering result~\cite{spectral} utilizing the self-attention map, the $3$-rd column shows the cross-attention map, the $4$-st column displays the attention score attained after employing the clustering technique in CBP~\cite{Mix-and-Match}, and the last column shows our final correlation map after propagation. 
  Best viewed in color.}
  \label{fig:prop}
\end{figure}
Previous works have considered exploiting cross-attention and self-attention layers of text-to-image diffusion models for localization tasks~\cite{Mix-and-Match, p2p}.
However, they often apply these two attention layers separately without fully leveraging their internal correlations.
We instead treat self-attention as the affinity matrix of different patches~\cite{affinitynet}, which conforms more to its essence compared to regarding it as the clustering feature~\cite{Mix-and-Match}, as shown in Figure~\ref{fig:prop}.

Specifically, the text condition $y$ is first projected to the CLIP embedding space as $h \in \mathbb{R}^{l \times d_t}$, where $l$ is the length of tokens and $d_t$ is its dimension.
The spatial visual feature at the intermediate layer $n$ of the backbone is denoted by $f^{(n)} \in \mathbb{R}^{WH^{(n)} \times d_i^{(n)}}$, where $WH^{(n)}$ represents the resolution and $d_i^{(n)}$ is its dimension.
$f^{(n)}$ interacts with $h$ via a cross-attention mechanism, which can derive a patch-token correlation map $Cross^{(n)} \in \mathbb{R}^{WH^{(n)} \times l}$ as:
\begin{equation}
    Cross^{(n)}=\operatorname{softmax}\left(\frac{Q_{cross}^{(n)} K_{cross}^{(n)T}}{\sqrt{d}}\right),
    \label{eq:softmax}
\end{equation}
where $Q_{cross}^{(n)} = f^{(n)} \cdot W_{Qc}^{(n)}$, $K_{cross}^{(n)} = h \cdot W_{Kc}^{(n)}$.
Here $W_{Qc}^{(n)} \in \mathbb{R}^{d_i^{(n)} \times d}$ and $W_{Kc}^{(n)} \in \mathbb{R}^{d_t \times d}$ are learned projection matrices while $d$ is the projection dimension.
Following a similar calculation process, we build the self-attention map as $Self^{(n)}=\operatorname{softmax}\left(\frac{Q_{self}^{(n)} K_{self}^{(n)T}}{\sqrt{d}}\right)$, with $Q_{self}^{(n)} = f^{(n)} \cdot W_{Qs}^{(n)}$, $K_{self}^{(n)} = f^{(n)} \cdot W_{Ks}^{(n)}$.
Here $W_{Qs}^{(n)} \in \mathbb{R}^{d_i^{(n)} \times d}$ and $W_{Ks}^{(n)} \in \mathbb{R}^{d_i^{(n)} \times d}$ are also learned projection matrices.

Inspired by~\cite{affinitynet, irnet}, we regard self-attention maps as semantic affinity matrices and propagate cross-attention scores accordingly.
Thereby, we aggregate semantic information from regions with similar appearances.
We select a particular layer with a fine resolution $WH$ to extract the self-attention map as $Self \in \mathbb{R}^{WH \times WH}$ and accumulate $Cross^{(n)}$ of different layers via interpolation and averaging to obtain the aggregated cross-attention map $Cross \in \mathbb{R}^{WH \times l}$.
We then leverage the patch-patch affinity matrix to refine the patch-token correlation map, which can be formulated as:
\begin{equation}
    SelfCross = Self^{iter} \cdot Cross,
    \label{eq:randomwalk}
\end{equation}
where $iter$ is the number of refining iterations. 
The obtained $SelfCross \in \mathbb{R}^{WH \times l}$ can serve as a substitute for $Cross$ to complete localization tasks~\cite{p2p}.
The originally predicted correlation between patch and token probably carries noise, yet ensembling within relevant areas could yield more stable results.
This refinement operation also declines forecast discrepancy within semantically similar regions.
As shown in Figure~\ref{fig:prop}, the correlation map attained after propagation preserves the structure better and possesses finer details.

To attain the pseudo mask, we leverage the image-level label to identify the categories contained in the image and use $K$ to represent this set.
Regarding class $k\in K$, we extract correlation maps of its relevant tokens from corresponding positions of $SelfCross$ and average along the token dimension to obtain the object attribution map $SC_k \in \mathbb{R}^{WH \times 1}$.
We reshape it to $\mathbb{R}^{W \times H}$ and then normalize it so that values at each position are between $0$ and $1$:
\begin{equation}
    SC_k(x, y) \rightarrow \frac{SC_k(x, y) - \min_{x, y} SC_k(x, y)}{\max_{x, y} SC_k(x, y) - \min_{x, y} SC_k(x, y)}.
    \label{eq:normalize}
\end{equation}
We further estimate the attribution map of background given by $SC_{bg}(x, y) = (1.0-\max_{k \in K} SC_k(x, y))^2$.
By concatenating these attribution maps, we acquire the final attention map $SC \in \mathbb{R}^{(|K|+1) \times W \times H}$.
After upsampling and refining $SC$, we assign the label for each pixel to the class (including the background) with the maximum attention score here.

\subsection{Text Query}
\label{sec:query}

When we want to locate specific objects in the image, we should provide a text query $y$ to distill the attention value.
Regarding the weakly-supervised semantic segmentation setting, we merge the category names contained in the image into one sentence.
For instance, if one image involves bottles, chairs, and a sofa, the text prompt will be \emph{``A photo including bottle, chair, and sofa."}.
This approach has several advantages.
First, it is more time-efficient than querying each object separately, as only one forward pass is required.
Second, the attention value of different objects will be comparable due to the normalization operation along the token dimension in Equation~\ref{eq:softmax}.
Last, we can conveniently append background prompts at the end of the sentence, which can alleviate the false-activation issue found in~\cite{clims}.
Moreover, class names in classic datasets may not fully represent the semantics of the category.
Therefore, we conduct prompt engineering to find synonyms with richer semantics and replace original category names with them.

\subsection{Summary}
\label{sec:summary}

We summarize our plug-and-play method by describing how one image is processed.
Given one image $x$, we first encode it using $\mathcal{E}$ to get the latent $z$.
We then obtain the noisy latent $z_t=\sqrt{\bar{\alpha}_t} z+\sqrt{1-\bar{\alpha}_t} \epsilon$ with a specific time step $t$ as described in Section~\ref{sec:preliminary} and input it into the denoising network $\epsilon_\theta$ along with the text query $y$ constructed as described in Section~\ref{sec:query}.
During the feed-forward calculation of $\epsilon_\theta(z_t, t, y)$, we distill $Cross$ and $Self$ from certain layers of $\epsilon_\theta$ and acquire $SelfCross$ using Equaion~\ref{eq:randomwalk}.
Next, we extract the attribution map of each entity from corresponding positions of $SelfCross$ and normalize it using Equaion~\ref{eq:normalize}.
Finally, after estimating the attribution map of background, we determine which entity each pixel belongs to by comparing the values of it on all attribution maps.

\section{Mug19 Dataset}
\label{sec:dataset}

\subsection{Motivation}

To investigate the problem of personalized segmentation, we create a dataset in a laboratory scenario.
As mentioned in Section~\ref{sec:intro}, this new task aims to locate the user-specific instance corresponding to the textual query.
It requires a more refined multi-modal comprehension capability than composed image retrieval~\cite{CIR} and may have potential application perspectives in products like home robotics.
Most existing datasets for similar tasks are designed for retrieval only~\cite{CIR, CIRR}, thus are not suitable for us.
\cite{PALAVRA} created a personalized segmentation benchmark repurposing from a video dataset~\cite{YouTube-VOS}.
However, the temporal continuity in the video may have information leakage for localization.
Besides, the depictions were not always useful for segmentation as we found it hard to tell the difference between similar entities in most images of this dataset.

\subsection{Distractors}

\begin{figure}
  \centering
  \includegraphics[width=\linewidth]{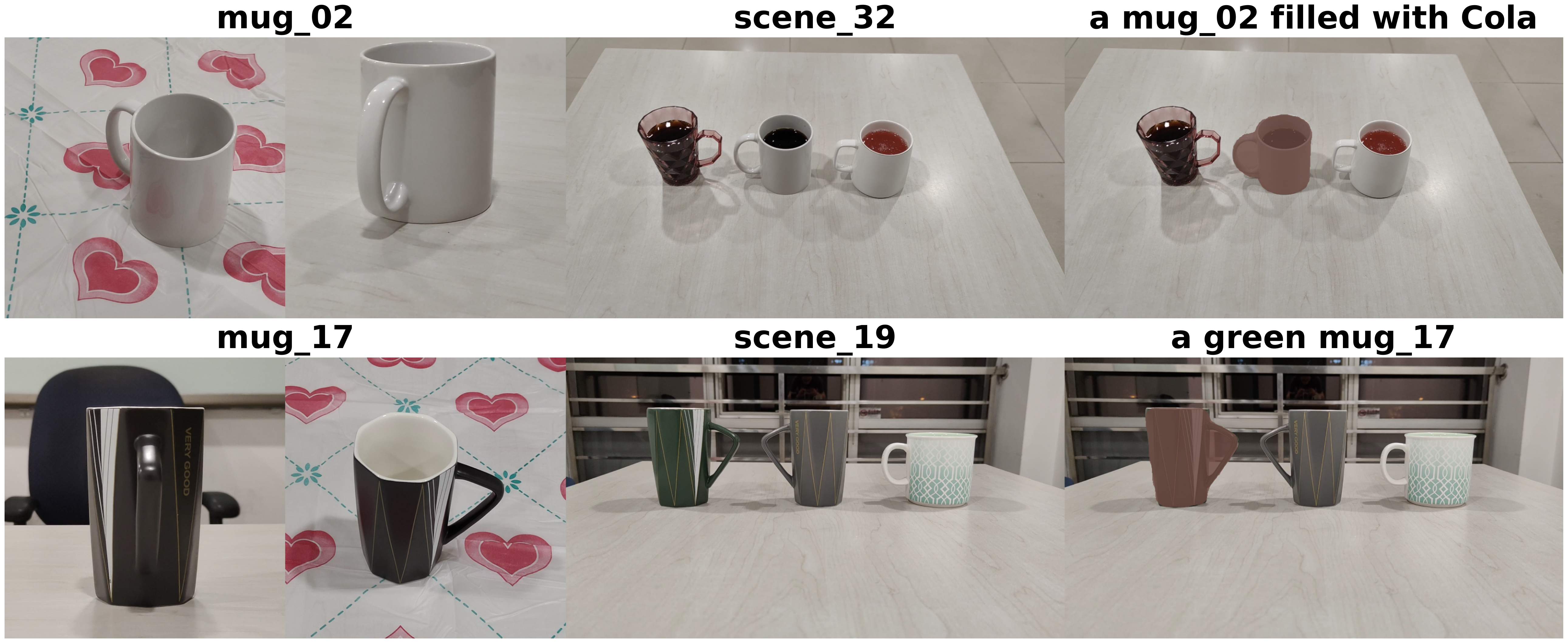}
  \caption{
  Examples in our proposed dataset.
  The first $2$ columns display multi-view photos of personalized items and the $3$-rd column presents the image of different scenes.
  The last column shows the highlighted segmentation result along with the text query.
  }
  \label{fig:dataset}
\end{figure}
To mitigate these issues, we build a new dataset on which models with good image-text reasoning ability would perform better.
We consider various ambiguities and ensure that the multi-modal references are necessary and sufficient to discriminate different objects, which requires a careful arrangement of scenes.
As shown in the $1$-st row of Figure~\ref{fig:dataset}, personalized knowledge is needed to tell the difference between $2$ mugs filled with Cola, while textual context is required to distinguish between $2$ white mugs.
These two situations are termed ``semantic distractors" and ``visual distractors" in~\cite{PALAVRA}.

\subsection{Statistics}

We choose a typical daily essential ``mug" to construct the dataset.
The dataset includes $19$ mugs, and they compose $47$ diverse scenes. 
We provide $5$ to $10$ $640 \times 640$ RGB images for each object. 
Each scene is consisted of $3$ to $5$ mugs and contains $100$ $1137 \times 640$ RGB images captured from different angles of view.
We create $2468$ segmentation triplets (instances in the scene, descriptions, scene image) based on our data, and we pick out $2$ special groups regarding the degree of ambiguity: semantic distractor and visual distractor.
Scenes in the semantic distractor split include different items with similar contexts and it consists of $440$ triplets.
Scenes in the visual distractor split contain various items with analogous appearances and this group comprises $170$ triplets.
The triplets are carefully annotated by experts and the instance mask labels are acquired by the pre-trained Mask R-CNN~\cite{maskrcnn} model in the Detectron2 library\footnote{\url{https://github.com/facebookresearch/detectron2}}.
This dataset will be made available online upon acceptance.

\section{Experiments}
\label{sec:experiment}

We evaluate our framework quantitatively and qualitatively on different segmentation tasks.
We further compare with strong baselines to showcase our strengths.
Lastly, we conduct ablation studies to analyze various designs.
The source code and new dataset are available at \url{https://github.com/Big-Brother-Pikachu/Text2Mask}.

\subsection{Weakly-Supervised Semantic Segmentation}
\label{sec:wsss}

\paragraph{Implementation Details}

We adopt Stable Diffusion for our experiment.
When processing one test image, we first rescale it to meet the model requirements and then encode it through $\mathcal{E}$.
We add noise to the latent and input it into the denoising network $\epsilon_\theta$.
The conditional language prompt $y$ is constructed as described in Section~\ref{sec:query}.
We choose the time step $t$ of the input noisy latent $z_t$ to be $150$ with the total denoising length $T=1000$.
With one feed-forward computation, we aggregate $Cross$ from the $4, 5, 6, 7, 8$-th layers of $\epsilon_\theta$ and distill $Self$ from the $11$-th layer of $\epsilon_\theta$.
We set $iter$ in Equaion~\ref{eq:randomwalk} to be $2$.
Following~\cite{clims, CLIP-ES}, the obtained segmentation maps are further refined by dense Conditional Random Fields (CRF)~\cite{dCRF} to generate pseudo masks, which are then used to train a standard segmentation network based on DeepLab~\cite{DeepLab}.
Furthermore, due to the stochastic sampling process in the diffusion framework, we can easily ensemble segmentation masks generated from different noises.
Thus we can sample multiple times for better pseudo masks.

\paragraph{Computation Consumption}

Our method requires no re-training and has $1066.2$M frozen parameters in total.
It takes an average of $0.88$ seconds to infer an image with a single NVIDIA GeForce RTX 3090 GPU and $24$ GB of memory.

\paragraph{Evaluation}

We select Pascal VOC 2012~\cite{pascalvoc} and Microsoft COCO 2014~\cite{mscoco} for evaluation. 
Pascal VOC 2012 is a semantic segmentation dataset with $20$ object categories.
It consists of $1464$ training images, $10582$ augmented training images, $1449$ validation images, and $1456$ test images.
MS COCO 2014 dataset contains $80$ object categories and $1$ background class.
It has $82081$ training images and $40137$ validation images.
In our experiment, only image-level labels are used to generate pseudo masks.
The mean Intersection over Union (mIoU) value is reported as the evaluation metric.

\paragraph{Quantitative Results}

\begin{table}
  \centering
  \resizebox{\textwidth}{!}{
  \begin{tabular}{lcccccccccccc}
    \hline
    Method  & 
    \begin{tabular}[c]{@{}c@{}}IRN\\\cite{irnet}\end{tabular} & 
    \begin{tabular}[c]{@{}c@{}}M\&L\\\cite{Li2023WeaklySS}\end{tabular} & \begin{tabular}[c]{@{}c@{}}SC-CAM\\\cite{SC-CAM}\end{tabular} & 
    \begin{tabular}[c]{@{}c@{}}SEAM\\\cite{SEAM}\end{tabular} & 
    \begin{tabular}[c]{@{}c@{}}AdvCAM\\\cite{AdvCAM}\end{tabular} & 
    \begin{tabular}[c]{@{}c@{}}CLIMS\\\cite{clims}\end{tabular} & 
    \begin{tabular}[c]{@{}c@{}}RIB\\\cite{RIB}\end{tabular} &
    \begin{tabular}[c]{@{}c@{}}OoD\\\cite{OoD}\end{tabular} &
    \begin{tabular}[c]{@{}c@{}}MCT\\\cite{MCTfomer}\end{tabular} &
    \begin{tabular}[c]{@{}c@{}}CLIP-ES\\\cite{CLIP-ES}\end{tabular} &
    \begin{tabular}[c]{@{}c@{}}Ours\\1 time\end{tabular} &
    \begin{tabular}[c]{@{}c@{}}Ours\\10 times\end{tabular} \\
    \hline
    Initial & 48.8 & 49.6 & 50.9 & 55.4 & 55.6 & 56.6 & 56.5 & 59.1 & 61.7 & 70.8 & 72.7 & \textbf{74.2} \\
    dCRF    & 54.3 &   -  & 55.3 & 56.8 & 62.1 & 62.4 & 62.9 & 65.5 & 64.5 & 75.0 & 74.4 & \textbf{76.1} \\
    RW      & 66.3 & 67.0 & 63.4 & 63.6 & 68.0 & 70.5 & 70.6 & 72.1 & 69.1 & - & - & - \\
    \hline
  \end{tabular}
  }
  \caption{mIoU of pseudo masks on PASCAL VOC 2012 \emph{train} set. The best results are in \textbf{bold}. dCRF represents post-processing with dense CRF. RW denotes refining with trained affinity networks.}
  \label{tab:pseudo_label}
\end{table}
\begin{table}
  \centering
  \begin{tabular}{lcccc}
    \hline
    Dataset & \multicolumn{2}{c}{VOC 2012 \emph{trainaug}} &  \multicolumn{2}{c}{COCO 2014 \emph{train}} \\
    \hline
    Method & CLIP-ES~\cite{CLIP-ES} & Ours & CLIP-ES~\cite{CLIP-ES} & Ours \\
    \hline
    Initial & 65.9 & \textbf{70.3} & 39.7 & \textbf{43.7} \\
    dCRF  & 68.7 & \textbf{71.7} & 41.5 & \textbf{45.3} \\
    \hline
  \end{tabular}
  \caption{mIoU of pseudo masks on the data set used to train DeepLab. 
  The best results are in \textbf{bold}.
  dCRF represents post-processing with dense CRF.}
  \label{tab:train_aug}
\end{table}
\begin{table}
  \centering
  \begin{tabular}{lclcc}
    \toprule
    Method  & Backbone   & Ver.    & Val   & Test \\
    \midrule
    \multicolumn{5}{l}{\textbf{Image-level supervision only.}} \\
    
    PSA~\cite{affinitynet}       & WR38       & V1               & 61.7  & 63.7 \\
    IRN~\cite{irnet}       & R50        & V2               & 63.5  & 64.8 \\
    ICD~\cite{ICD}       & R101       & V1$^\ddagger$    & 64.1  & 64.3 \\
    SC-CAM~\cite{SC-CAM}    & R101       & V2$^\ddagger$    & 66.1  & 65.9 \\
    BES~\cite{BES}       & R101       & V2$^\ddagger$    & 65.7  & 66.6 \\
    ETG~\cite{EraseTG} & R101       & V2    & 66.8 & 67.6 \\
    M\&L~\cite{Li2023WeaklySS} & R101       & V2    & 67.2 & 69.1 \\
    SIPE~\cite{SIPE}      & R101       & V2$^\ddagger$    & 68.8  & 69.7 \\
    RIB~\cite{RIB}       & R101       & V2               & 68.3  & 68.6 \\
    AMN~\cite{AMN}       & R101       & V2$^\ddagger$    & 70.7  & 70.6 \\
    \midrule
    \multicolumn{5}{l}{\textbf{Image-level supervision + Language supervision.}} \\
    
    CLIMS~\cite{clims}     & R101       & V2               & 69.3  & 68.7 \\
    CLIMS~\cite{clims}     & R101       & V2$^\ddagger$    & 70.4  & 70.0 \\
    CLIP-ES~\cite{CLIP-ES}   & R101  & V2               & 71.1 & 71.4 \\
    CLIP-ES~\cite{CLIP-ES}   & R101  & V2$^\ddagger$    & \textbf{73.8} & 73.9 \\
    Ours   & R101  & V2               & \textbf{71.2} & \textbf{71.5} \\
    Ours   & R101  & V2$^\ddagger$    & 73.3 & \textbf{74.2} \\
    \bottomrule
  \end{tabular}
  \caption{
  DeepLab results on PASCAL VOC 2012 \emph{val} and \emph{test} sets. 
  The best results are in \textbf{bold}. 
  Ver. denotes the version. 
  $^\ddagger$ represents adopting COCO pre-trained models.}
  \label{tab:deeplab}
\end{table}
\begin{table}
  \centering
  \begin{tabular}{lclc}
    \toprule
    Method  & Backbone   & Sup.    & Val \\
    \midrule
    ETG~\cite{EraseTG}       & R101        & I    & 28.0 \\
    IRN~\cite{irnet}       & R50        & I    & 32.6 \\
    IRN~\cite{irnet}       & R101       & I    & 41.4 \\
    SIPE~\cite{SIPE}      & R101       & I    & 40.6 \\
    RIB~\cite{RIB}       & R101       & I    & 43.8 \\
    AMN~\cite{AMN}       & R101       & I    & 44.7 \\
    CLIP-ES~\cite{CLIP-ES} & R101       & I+L    & 45.4 \\
    Ours                    & R101       & I+L    & \textbf{45.7} \\
    \bottomrule
  \end{tabular}
  \caption{
  DeepLab results on MS COCO 2014 \emph{val} set. 
  The best results are in \textbf{bold}. 
  I and L represents image-level supervision and language supervision, respectively.
  }
  \label{tab:COCO_val}
\end{table}
We first evaluate the quality of our pseudo masks on Pascal VOC 2012.
As shown in Table~\ref{tab:pseudo_label}, our method outperforms previous methods by a considerable margin on initial seeds.
Without training an extra affinity network, we can achieve $76.1\%$ mIoU on the \emph{train} set after dCRF post-processing.
To train the segmentation model, we generate pseudo masks on the \emph{trainaug} set.
As shown in Table~\ref{tab:train_aug}, we obtain the best quality on this set, and even our initial masks can exceed the results of previous methods post-processing with dCRF.
Utilizing these masks, we train segmentation models based on DeepLabV2 and assess them on the \emph{val} and \emph{test} sets.
As shown in Table~\ref{tab:deeplab}, our framework outperforms most previous methods and we achieve a new state-of-the-art with the Imagenet pre-trained model.
As for MS COCO 2014, we first compare the generated pseudo masks on the \emph{train} images.
As shown in Table~\ref{tab:train_aug}, our method produces more precise masks.
After training a DeepLabV2 segmentation model with these pseudo masks, we can achieve $45.7\%$ mIoU on the \emph{val} set as shown in Table~\ref{tab:COCO_val}, which is also a new SOTA.
As we adopt the classic DeepLab repository directly without extensive modifications, the enhancement observed in the DeepLab results is relatively modest.
However, we believe that investing excessive time in tuning the training process of DeepLab may not be cost-effective.
Our plug-and-play approach inherently holds the advantage of simplicity over most competing methods, as we don't require extra optimization processes as in CLIMS~\cite{clims}.
Besides, generative models are capable of determining where to locate an entity while discriminative models can only ascertain its existence.
Thus we perform better than the discriminative-model-based method CLIP-ES~\cite{CLIP-ES}.

\paragraph{Qualitative Results}

\begin{figure}
  \centering
  \includegraphics[width=\linewidth]{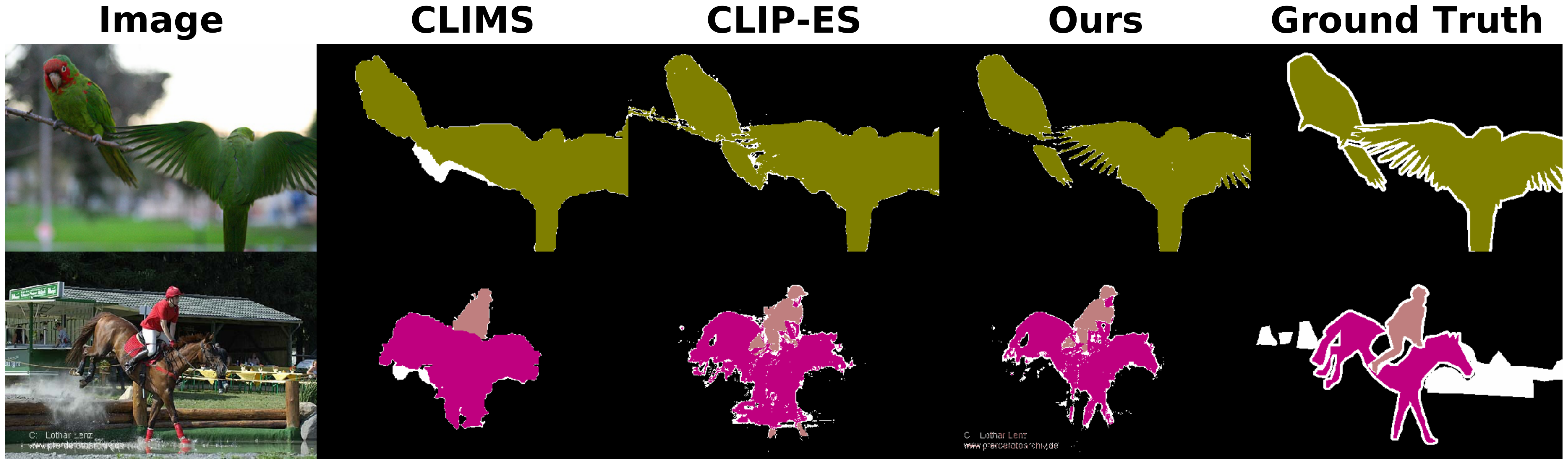}
  \caption{
  Visualizations of the pseudo masks generated by various methods.
  The $1$-st column shows the input image and the last column shows the ground truth mask.
  Uncertain pixels are set to white.
  }
  \label{fig:compare}
\end{figure}
We visualize the results of our approach and other multi-modal related methods~\cite{clims, CLIP-ES} in Figure~\ref{fig:compare}.
We produce pseudo masks with finer structures, such as the feather of the bird and the legs of the horse as shown in the image.
Furthermore, we can directly generate the confidence map from attention scores.
If the value of a position is within the $0.05$ interval of the interface, we set it as uncertain.
We find that ambiguous pixels are mainly concentrated on object boundaries.
More results are shown in Appendix.

\subsection{Ablation Study}
\label{sec:ablation}

\paragraph{Attention}

\begin{table}
  \centering
  \begin{tabular}{ccccc}
    \toprule
    Attention  & $Cross$ & CBP~\cite{Mix-and-Match} & $SelfCross$ & $Cross$+$SelfCross$ \\
    \midrule
    Initial & 61.7 & 64.1 & \textbf{72.7} & 69.9 \\
    dCRF & 67.1 & 69.9 & \textbf{74.4} & 73.0 \\
    \bottomrule
  \end{tabular}
  \caption{Attention mechanism analysis on PASCAL VOC 2012 \emph{train} set. The best results are in \textbf{bold}.}
  \label{tab:attention}
\end{table}
We verify the effectiveness of our attention mechanism based on the quality of pseudo masks.
As shown in Table~\ref{tab:attention}, $SelfCross$ achieves remarkably better mIoU than $Cross$, and a qualitative comparison has been displayed in Figure~\ref{fig:prop}.
A further ensemble of $Cross$ and $SelfCross$ shows no benefits.
We also compare to Controllable Background Preservation (CBP) from~\cite{Mix-and-Match}, which clusters the self-attention map and assigns labels to each segment based on the cross-attention value.
We achieve better results as our approach conforms more to the essence of these two attention maps.

\paragraph{Prompt}

\begin{table}
  \centering
  \begin{tabular}{ccc|ccccc|c}
    \hline
    Comp. & Syn. & BG. & bird & boat & person & train & tvmonitor & mIoU\\
    \hline
      &   &   & 85.6 & \textbf{74.2} & 38.4 & 74.1 & 62.3 & 72.1 \\
      & \checkmark &   & \textbf{87.1} & 72.8 & 59.3 & 73.5 & 60.6 & 73.1 \\
      &   & \checkmark & 80.3 & 71.7 & 36.1 & \textbf{81.9} & 66.3 & 72.7 \\
      & \checkmark & \checkmark & 83.9 & 71.6 & \textbf{63.5} & 81.5 & \textbf{67.3} & 74.3 \\
    \cline{1-8}
    \checkmark &   &   & 85.3 & 76.4 & 45.2 & 75.2 & 62.8 & 73.2 \\
    \checkmark & \checkmark &   & \textbf{87.0} & 75.7 & 56.9 & 74.7 & 64.5 & 73.8 \\
    \checkmark &   & \checkmark & 79.4 & \textbf{77.1} & 41.0 & 82.2 & \textbf{66.3} & 73.1 \\
    \checkmark & \checkmark & \checkmark & 85.0 & 77.0 & \textbf{59.7} & \textbf{82.5} & 65.6 & \textbf{74.4} \\
    \hline
  \end{tabular}
  \caption{
  Prompt strategy comparison on PASCAL VOC 2012 \emph{train} set.
  Comp. denotes using one composed sentence, Syn. denotes adopting category synonyms, and BG. denotes appending background prompts.
  The best results are in \textbf{bold}.}
  \label{tab:prompt}
\end{table}
We analyze different strategies for selecting the text query and depict the IoU of some categories along with the overall average in Table~\ref{tab:prompt}.
Compared to using one composed sentence as the prompt, applying separate texts to query multi-label scenes will decline segmentation precision and increase inference time (the number of pseudo labels generated per second decreases from $1.6$ to $1.0$ in practice).
Then following~\cite{CLIP-ES}, we substitute the original class name with category synonyms that can reduce ambiguity.
For instance, the high-attention region corresponding to \emph{``person"} tends to overlook the body, and adding \emph{``clothes"} to the query prompt of \emph{``person"} can enhance the performance.
Furthermore, we append some background prompts at the end of text queries.
With the existing $\operatorname{softmax}$ operation in Equation~\ref{eq:softmax}, category scores in co-occurring background regions~\cite{clims} will be naturally suppressed.
For example, when \emph{``railway"} and \emph{``track"} are used as the background prompts for \emph{``train"}, the segmentation results will be improved due to the exclusion of these background areas.

\subsection{Personalized Referring Image Segmentation}
\label{sec:personalized}

\paragraph{Implementation Details}

This task aims to locate personalized instances in the scene image with the help of textual descriptions.
We first leverage Custom Diffusion~\cite{customdiffusion} to acquire the text embedding of the personalized item.
Given target object images, we optimize the token embedding of the identifier word \emph{``$<$new1$>$"} along with the diffusion model.
Next, we replace \emph{``$<$class$>$"} in the textual context with \emph{``$<$new1$>$ $<$class$>$"} as the condition text prompt, and then attain the object attribution map from $SelfCross$ using the same hyper-parameters as in Section~\ref{sec:wsss}.
To focus on the foreground region, we query the diffusion model with \emph{``A photo including $<$class$>$."} first to get the object area in the scene.
Then we compare the attribution map of different instances to determine the segmentation results in the object area.
We have found it hard to gain a complete silhouette with the above process.
Therefore, we add a clustering process in advance, so that we only have to focus on assigning regions instead of pixels.
We employ a simple version of spectral clustering~\cite{spectral} which performs surprisingly well in separating each instance.
Later, we average the attention value in different segments and finally assign the instance to the area with the highest correlation.

\paragraph{Computation Consumption}

The training process of Custom Diffusion takes around $6$ minutes on $2$ A100 GPUs for each instance, and the inference cost is similar to that in Section~\ref{sec:wsss}.

\paragraph{Evaluation}

We use standard mIoU along with two kinds of accuracy for evaluation.
Accuracy represents the proportion of correctly predicted instances to all instances.
The ground truth assignment between instances and segments is determined based on the relative relationship between the center positions of segments.
When we consider each object solely, the segment most relevant to each object is predicted as its location.
$bf\_acc$ denotes the accuracy under this protocol.
When we consider all the items contained in the scene together, we treat the correlation between items and segments as the cost matrix of an assignment problem.
Then Hungarian algorithm~\cite{hungarian} is adopted to attain the assignment between instances and segments. 
$af\_acc$ denotes the accuracy under this protocol.
Usually, the latter accuracy will be higher as we compare the correlation values not only between segments but also between contained instances.

\paragraph{Baselines}

We compare our method with several strong baselines. 
First, we use the feature of Mask R-CNN~\cite{maskrcnn} to calculate the similarity between queries and instances in the scene.
It is worth noting that we have used the same model to provide instance masks and we aim to evaluate the ability of this localization model to discriminate similar objects.
We also adopt DINO-ViT, a self-supervised Vision Transformer model, as dense visual descriptors following~\cite{dino}.
Next, we simply query the image with \emph{``a $<$new1$>$ $<$class$>$"} and we call this baseline ``Subject Only".
On the contrary, the ``Context Only" baseline stands for not incorporating the identifier embedding in the textual prompt.
These approaches only use uni-modal information.
Then the ``Arithmetic" baseline combines subject and context by replacing the category embedding in the description with the average between CLIP image embeddings of the instance and the CLIP text embedding of the class.

\paragraph{Quantitative Results}

\begin{table}
  \centering
  \setlength\tabcolsep{0.95pt}
  \begin{tabular}{l|ccc|ccc|ccc}
    \hline
    Split  & \multicolumn{3}{c|}{All} & \multicolumn{3}{c|}{Semantic Distractor} & \multicolumn{3}{c}{Visual Distractor} \\
    \hline
    Metric  & mIoU & bf\_acc & af\_acc & mIoU & bf\_acc & af\_acc & mIoU & bf\_acc & af\_acc  \\
    \hline
    Mask R-CNN~\cite{maskrcnn}       & \textbf{75.1} & 58.9 & 75.1 & \underline{78.2} & 63.3 & 78.2 & 42.9 & 38.0 & 42.9 \\
    DINO-ViT~\cite{dino}       & 52.7 & 42.3 & 72.8 & 70.7 & 43.3 & 93.2 & 39.9 & 33.7 & 52.0 \\
    Subject Only       & 62.3 & \underline{60.1} & \underline{79.5} & 76.2 & \textbf{79.3} & \underline{95.0} & 31.9 & 42.0 & 40.0 \\
    Context Only      & 54.4 & 51.2 & 70.9 & 35.3 & 40.5 & 45.1 & 50.3 & 42.4 & 64.3 \\
    Arithmetic & 28.5 & 35.1 & 37.2 & 27.9 & 37.7 & 35.2 & \underline{55.2} & \underline{46.7} & \underline{70.4} \\
    Ours & \underline{64.9} & \textbf{60.2} & \textbf{83.3} & \textbf{78.8} & \underline{73.5} & \textbf{98.3} & \textbf{56.8} & \textbf{49.4} & \textbf{71.8} \\
    \hline
  \end{tabular}
  \caption{Evaluation results on different splits of \textbf{Mug19} dataset. The best results are in \textbf{bold} while the second best are \underline{underlined}.}
  \label{tab:personalized}
\end{table}
As shown in Table~\ref{tab:personalized}, our method achieves the best results on the entire data.
Mask R-CNN obtains good mIoU because it knows the exact position of all mugs in advance, but it performs poorly in distinguishing between various mugs.
It is also found that the subject information is beneficial for excluding semantic distractors while the context information is useful for dealing with visual distractors, which is in line with our definition of these two groups.
Besides, the average operation can not fully leverage the information from different modalities, as the ``Arithmetic" baseline behaves similarly to ``Context Only" on various splits.
We can also conclude that multi-modal knowledge is vital for this task, as uni-modal methods have difficulties in handling distinct distractors.

\paragraph{Case Study}

\begin{figure}
  \centering
  \includegraphics[width=\textwidth]{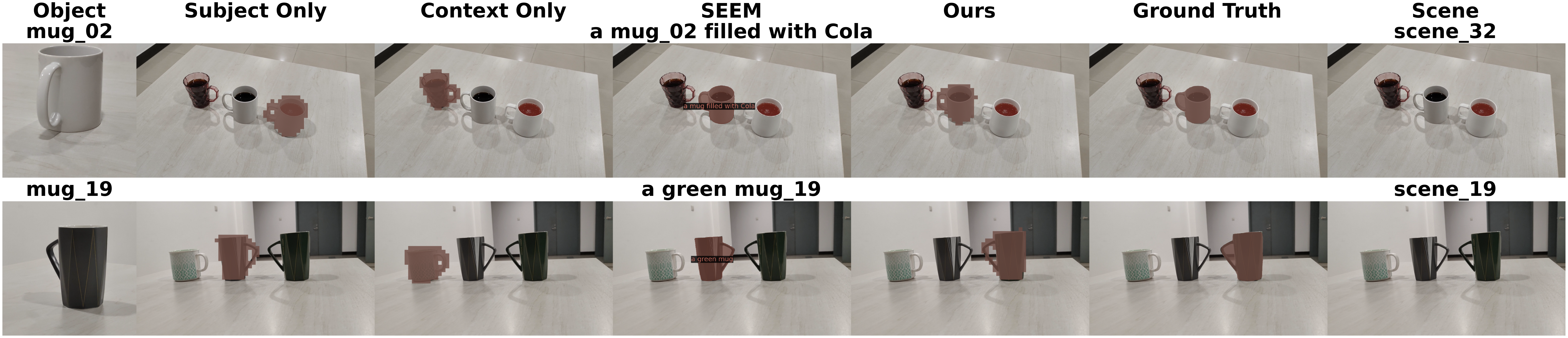}
  \caption{
  Localization results of different methods on \textbf{Mug19} dataset samples.
  The $1$-st column shows the object, the last column shows the scene and the rest columns display highlighted segmentation masks with the text reference.
  }
  \label{fig:case}
\end{figure}
We select several samples to showcase the deviation of different approaches.
As shown in Figure~\ref{fig:case}, the ``Subject Only" baseline has difficulties in distinguishing visual distractors while the ``Context Only" baseline is hard to discriminate semantic distractors.
More specifically, the former one finds another mug with a similar appearance in the $1$-st row, and the latter baseline detects another \emph{``green"} mug in the $2$-nd row.
We also compare our method with the recent work SEEM~\cite{SEEM}.
We use its official demo and simultaneously choose example and text interactive mode.
The result of the $2$-nd case indicates that it sometimes neglects textual information.
Compared to these methods, our approach leverages multi-modal information more efficiently.

\section{Limitations and Discussions}
\label{sec:limitation}

\begin{figure}
  \centering
  \includegraphics[width=\linewidth]{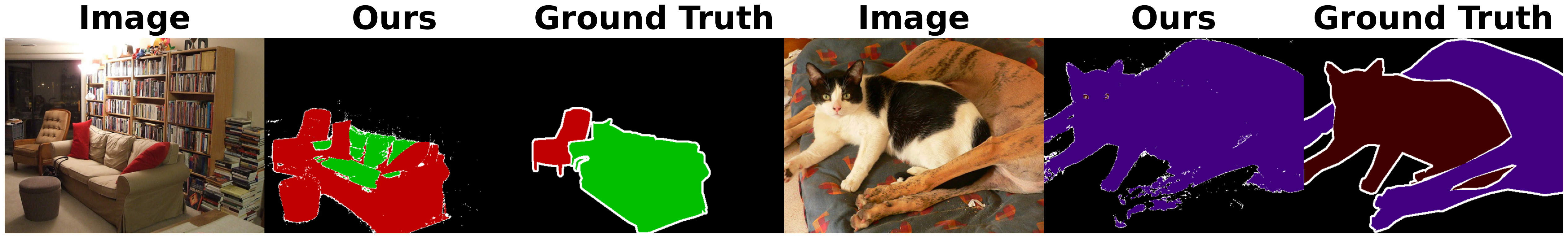}
  \caption{
  Failure cases: when \emph{``chair"} and \emph{``sofa"}, \emph{``cat"} and \emph{``dog"} are in the same image, it is difficult for our method to distinguish them.
  }
  \label{fig:cohyponym}
\end{figure}
We have found that the segmentation results will drop drastically when an image contains semantically similar objects as shown in Figure~\ref{fig:cohyponym}.
The same phenomenon has also been found in~\cite{daam} and was termed ``Cohyponym Entanglement".
Incorporating extra visual knowledge into the learning stage of diffusion models~\cite{ERNIE} may also help enhance their discriminative abilities.
Besides, we have made preliminary attempts to adopt affordance-related texts.
When considering texts with richer semantics like \emph{``graspable part"} and \emph{``cuttable part"}, our framework can provide meaningful masks for images containing knives.
We will further explore how to efficiently equip our framework with affordance ability in future works.

\section{Broader Impacts}
\label{sec:impacts}

The booming generative models have caused many concerns.
The unauthentic content they create can be misused.
We instead pay attention to the discriminative task and show the potential of these generative models from another aspect.
Our method can further help researchers to understand the generation process in large models, which will also benefit the detection of generated content.
On the other hand, as we propose a more flexible segmentation technology, surveillance of specific identities becomes easier, which may be abused to infringe on personal privacy.

\section{Conclusion}
\label{sec:conclusion}

In this work, we introduce a simple but effective approach to distill the attention value in text-to-image diffusion models for segmentation.
Without re-training nor inference-time optimization, text-related regions can be located precisely.
We first validate the effectiveness of our framework on weakly-supervised semantic segmentation tasks.
To this end, we propose a novel way to generate query prompts in accordance with image-level labels.
Our framework achieves state-of-the-art performance on PASCAL VOC 2012 and MS COCO 2014 and the validity of our designs is confirmed.
To further verify the role of word-pixel correlation, we introduce a new practical task named ``personalized referring image segmentation" with a new real-world dataset.
Experiments on this task demonstrate that our method possesses a better multi-modal comprehension ability than several strong baselines.

\section*{Acknowledgments}
\label{acknowledgments}

This work is funded by the National Science and Technology Major Project of China (No. 2022ZD0114903).

\appendix

\section{Experiment Details}

 \paragraph{Datasets and Pre-trained Models}

 The licenses of the datasets and the pre-trained models are listed here.
 The PASCAL VOC 2012 dataset~\cite{pascalvoc} is from~\url{http://host.robots.ox.ac.uk/pascal/VOC/index.html}.
 The MS COCO 2014 dataset~\cite{mscoco} is from~\url{https://cocodataset.org/#home}.
 The pre-trained CLIP~\cite{clip} is from~\url{https://github.com/openai/CLIP}.
 The Stable Diffusion v$1.4$ model~\cite{stablediffusion} is from~\url{https://huggingface.co/CompVis/stable-diffusion}.
 The DeepLab pre-trained model~\cite{DeepLab} is from~\url{https://github.com/kazuto1011/deeplab-pytorch}.
 The Detectron2 pre-trained model is from~\url{https://github.com/facebookresearch/detectron2}, and we select mask\_rcnn\_R\_50\_FPN\_3x from official baseline models for its good performance.

 \paragraph{Implementation Details}

 We use Pytorch~\cite{pytorch} for our experiments. 
 Our codebase builds heavily on~\url{https://github.com/CompVis/stable-diffusion}.
 For Pascal VOC 2012, MS COCO 2014, and \textbf{Mug19} datasets, we sample $10$, $1$, and $1$ times respectively to generate pseudo masks.

 \paragraph{Prompts}

 We adopt the category synonyms in~\cite{CLIP-ES} except for \emph{``person"}, as we find \emph{``person with clothes"} works better than \emph{``person with clothes, people, human"}.
 We use \emph{``tree, river, sea, lake, water, railway, railroad, track, stone, rocks"} as the background prompts following~\cite{clims}.
 We find the background prompts in~\cite{CLIP-ES} work slightly worse.
 We attribute it to more background classes, which leads to more complex query sentences as we append background prompts to the end of texts.

 \paragraph{Training Details of DeepLab}

 We refer to the config file in \url{https://github.com/kazuto1011/deeplab-pytorch} and \url{https://github.com/CVI-SZU/CLIMS}.
 We adopt DeepLabV2 and use ResNet-101 as the backbone.

 \section{More Experiment Results}

 \subsection{Weakly-Supervised Semantic Segmentation}

 \paragraph{Qualitative Results}

 \begin{figure}
   \centering
   \includegraphics[width=\textwidth]{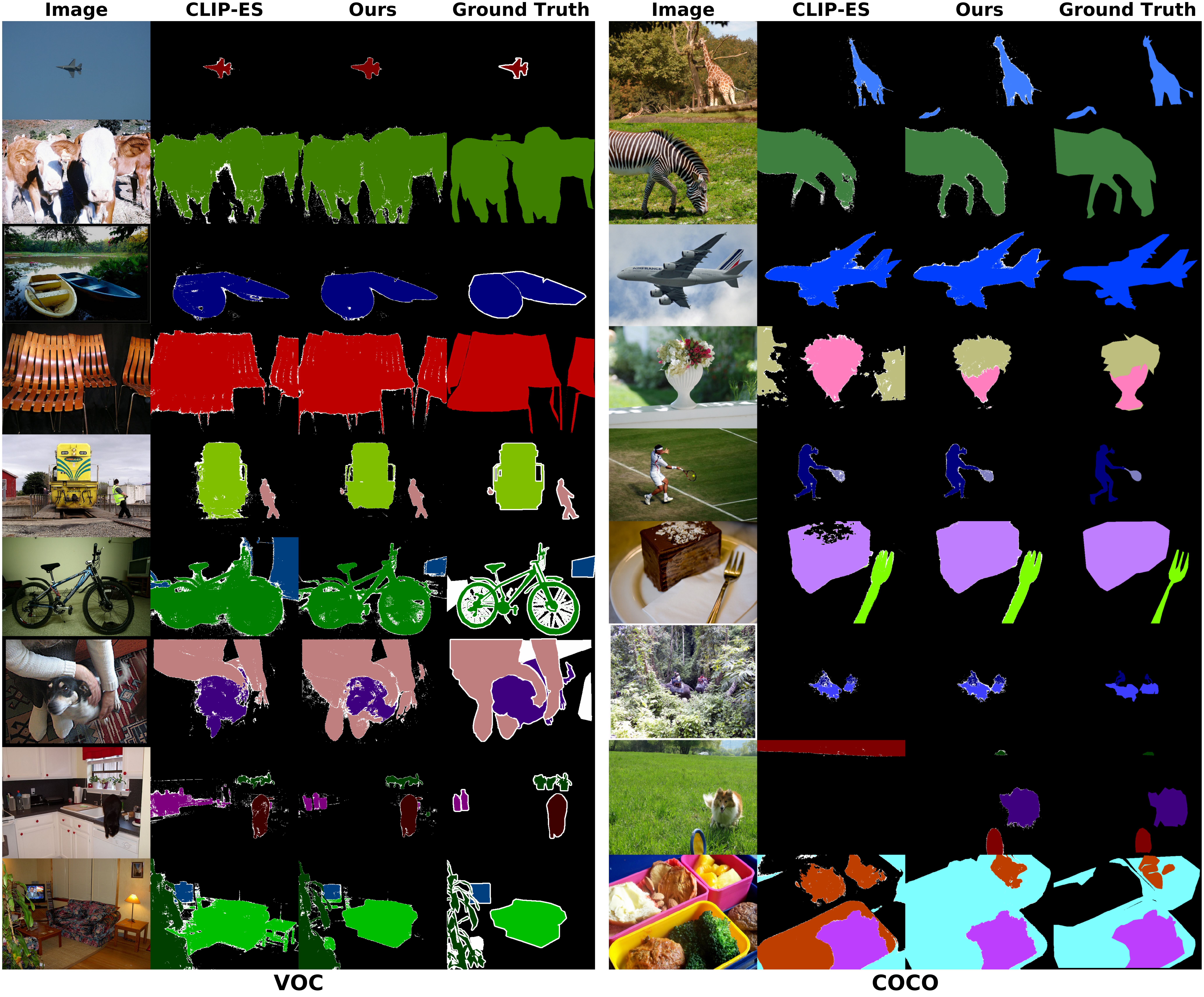}
   \caption{
   More visualization results on PASCAL VOC 2012 and MS COCO 2014 datasets.
   Uncertain pixels are set to white.
   }
   \label{fig:qualitative}
 \end{figure}
 We show more qualitative comparisons of our generated pseudo masks in Figure~\ref{fig:qualitative}.
 We can observe that our framework accurately locates objects of different sizes in both simple and complex scenarios.
 
  \paragraph{Hyper-parameters}

 \begin{figure}
   \centering
   \includegraphics[width=\linewidth]{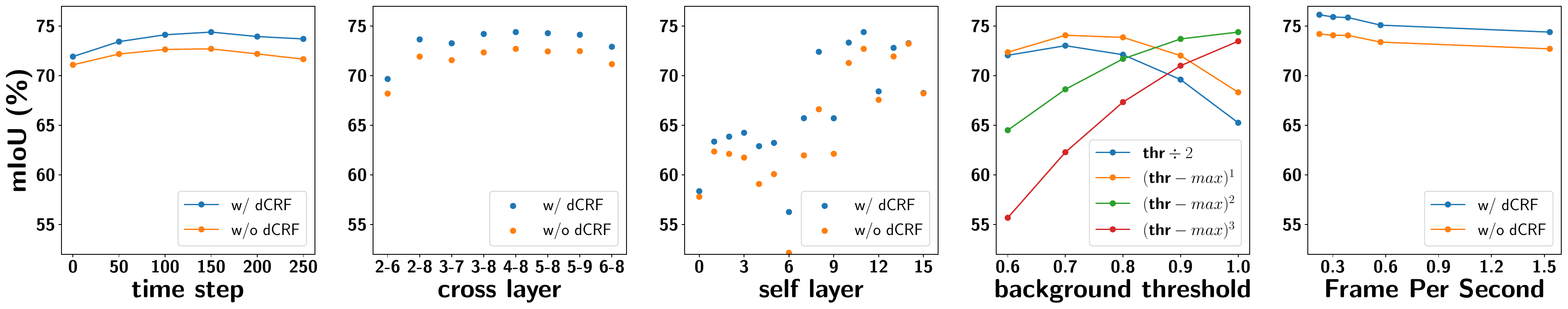}
   \caption{
   Analyses of different hyper-parameters.
   The mIoU values are evaluated on PASCAL VOC 2012 \emph{train} set.
   The default setting is $150$ time step, $4$-$8$-th cross layer, $11$-th self layer, $1.0$ background threshold with power $2$, and sampling $1$ time.
   }
   \label{fig:ablation}
 \end{figure}
 We analyze the impact of different hyper-parameters and compare the quality of generated pseudo masks in Figure~\ref{fig:ablation}.
 The time step $t$ of the input noisy latent $z_t$ has little effect on the result, while the performance is better when employing coarse inner layers for $Cross$ layers, where the text condition is more relevant to the layout according to a recent study~\cite{differentlayer}.
 On the contrary, higher $Self$ layers boost the performance more, since deeper layers tend to have a better understanding of appearance similarity.
 As for determining background regions, we have found a constant threshold less effective.
 We follow~\cite{CLIP-ES} with a minus-max-power mechanism where the background score per pixel is determined by threshold $\textbf{thr}$ and the $max$imum category attention value here.
 We compare different power exponents and square is the best.
 Lastly, as sampling more times requires more time, the trade-off between time and performance should be considered.
 We sample once for ablation studies and COCO experiments to save time, while for VOC, we sample $10$ times for better performance.

 \paragraph{Different Seeds}

 \begin{table}
   \centering
   \begin{tabular}{ccccccc}
     \toprule
     Seeds  & 40 & 41 & 42 & 43 & 44 & Average \\
     \midrule
     mIoU & 74.5 & 74.0 & 74.4 & 74.6 & 74.4 & $74.4 \pm 0.2$ \\
     \bottomrule
   \end{tabular}
   \caption{mIoU of pseudo masks on PASCAL VOC 2012 \emph{train} set with different seeds.}
   \label{tab:seed}
 \end{table}
 As we can obtain better performance when sampling several times, it would be interesting to probe the performance changes with different seeds.
 We run $5$ experiments to generate pseudo masks on PASCAL VOC 2012 \emph{train} set with different seeds, and we only sample once for each time. 
 As shown in Table~\ref{tab:seed}, the results are stable, which indicates that the improvements brought by the ensemble operation may come from its ability to make the results in uncertain regions more reliable.

 \subsection{Personalized Referring Image Segmentation}

 \paragraph{Combinations of Personalized and Localized Methods}

 \begin{table}
   \centering
   \scalebox{0.85}{
   \setlength\tabcolsep{0.95pt}
   \begin{tabular}{l|ccc|ccc|ccc}
     \hline
     Split  & \multicolumn{3}{c|}{All} & \multicolumn{3}{c|}{Semantic Distractor} & \multicolumn{3}{c}{Visual Distractor} \\
     \hline
     Metric  & mIoU & bf\_acc & af\_acc & mIoU & bf\_acc & af\_acc & mIoU & bf\_acc & af\_acc  \\
     \hline
     PALAVRA\cite{PALAVRA}+CLIP-ES\cite{CLIP-ES}      & 26.0 & 44.5 & 31.5 & 2.5 & 33.0 & 3.0 & 11.5 & 38.8 & 14.1 \\
     Custom\cite{customdiffusion}+CLIP-ES\cite{CLIP-ES}      & 36.1 & 37.6 & 43.0 & 33.4 & 37.1 & 39.6 & 53.4 & 42.0 & 63.5 \\
     PALAVRA\cite{PALAVRA}+$SelfCross$      & \underline{53.1} & \underline{48.2} & \underline{69.4} & \underline{53.6} & \underline{46.3} & \underline{68.3} & \textbf{58.2} & \textbf{50.8} & \textbf{74.5} \\
     Custom\cite{customdiffusion}+$SelfCross$ & \textbf{64.9} & \textbf{60.2} & \textbf{83.3} & \textbf{78.8} & \textbf{73.5} & \textbf{98.3} & \underline{56.8} & \underline{49.4} & \underline{71.8} \\
     \hline
   \end{tabular}
   }
   \caption{
   Evaluation results of various combinations on \textbf{Mug19} dataset. 
   The best results are in \textbf{bold} while the second best results are \underline{underlined}.
   Custom + $SelfCross$ is the combination we ultimately adopt.
   }
   \label{tab:compose}
 \end{table}
 We conduct a study concerning distinct combinations of personalized and localized approaches.
 We adopt discriminative model-based methods: PALAVRA~\cite{PALAVRA} for personalization, CLIP-ES~\cite{CLIP-ES} for segmentation, and generative model-based methods: Custom Diffusion~\cite{customdiffusion} for personalization, our framework for segmentation.
 As shown in Table~\ref{tab:compose}, CLIP-ES has difficulties dealing with personalized items.
 We attribute it to the instability of Grad-CAM~\cite{gradcam} as we occasionally get all zero CAMs for specific objects.
 In contrast, our mechanism is robust to various personalized embeddings and we ultimately choose Custom Diffusion because it performs better and is more consistent with our method.

 \paragraph{Objects in the Same Series}

 We further construct a new split from \textbf{Mug19} dataset regarding series.
 We build a scenario where we want to locate the object with only access to images of its variants within the same series. 
 For instance, we aim to find the red transparent plastic mug in the scene, but we only possess photos of its blue variant.
 We handle it by using the identifier embedding of the blue variant along with the context \emph{``a red mug"} as the segmentation query.
 We pick out $632$ triplets based on this notion and name them ``variant split" as a whole.
 Quantitative comparisons of various baselines are conducted on this split in Table~\ref{tab:variant}.
 Our method fully utilizes both subject and context information and addresses the localization issue of this situation more effectively than all the baselines.
 \begin{table}
   \centering
   \begin{tabular}{l|ccc}
     \hline
     Split  & \multicolumn{3}{c}{Variant} \\
     \hline
     Metric  & mIoU & bf\_acc & af\_acc \\
     \hline
     Mask R-CNN~\cite{maskrcnn}       & \textbf{71.9} & \underline{59.5} & 71.9 \\
     DINO-ViT~\cite{dino}       & 48.5 & 44.7 & 66.7 \\
     Subject Only       & 61.6 & 54.9 & \underline{78.8} \\
     Context Only      & 54.7 & 52.6 & 71.1 \\
     Arithmetic & 36.5 & 37.3 & 47.1 \\
     Ours & \underline{69.2} & \textbf{60.7} & \textbf{88.3} \\
     \hline
   \end{tabular}
   \caption{Evaluation results on the variant split of \textbf{Mug19} dataset. The best results are in \textbf{bold} while the second best results are \underline{underlined}.}
   \label{tab:variant}
 \end{table}

 \section{Comparisons with SAM}
 \label{sec:sam}

 Drawing inspiration from Large Language Models (LLM), foundation models for segmentation have been developed recently~\cite{SAM, SEEM}.
 They can solve universal segmentation tasks with prompts of various modalities.
 However, this series of methods need dense annotations while our framework only requires image-level labels.

 \section{More Discussions on Limitations}
 \label{sec:dataset limitation}

 \begin{figure}
   \centering
   \includegraphics[width=\linewidth]{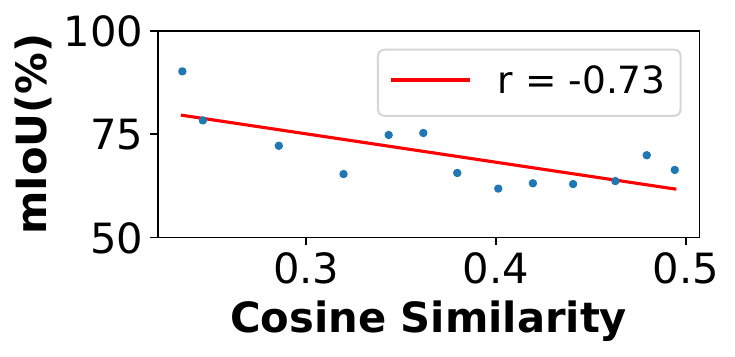}
   \caption{
   The correlation between segmentation results and textual embedding similarity.
   The r-value is -$0.73$.
   }
   \label{fig:coscim}
 \end{figure}
 First, we further analyze the situation where an image contains semantically similar objects.
 We compare the textual embeddings of different entities in one image, and choose images in VOC \emph{train} set that contain at least $2$ classes to calculate the results.
 For each image, we compute the average cosine similarity between different classes and record the mIoU.
 As shown in Figure~\ref{fig:coscim}, the more similar different entities in one image, the worse the segmentation results of our method, which is consistent with our finding in the \textbf{Limitations and Discussions} section.
 The p-value of rejecting the null hypothesis that the slope is zero is $0.0049$.
 Next, we discuss the limitations of our proposed task.
 We care more about the appearance of different personalized entities, and our descriptions currently focus more on what they look like.
 As CLIP, the text encoder of Stable Diffusion, struggles with understanding spatial relations~\cite{composerelation}, we doubt our framework's ability on them and have not yet considered them in our dataset.
 In future work, we will add this type of description to our dataset and consider how to enhance our method's ability to comprehend them.


\bibliographystyle{elsarticle-num-names}
\bibliography{elsarticle-template-num-names}





\end{document}